\documentclass{article}

\usepackage{spconf,amsmath,graphicx}
\usepackage{amsfonts}
\usepackage[flushleft]{threeparttable}
\usepackage{booktabs}
\usepackage{tabularx}
\usepackage{multirow}
\usepackage{subcaption}
\usepackage[table,dvipsnames]{xcolor}
\usepackage{amssymb}
\usepackage{url}
\usepackage{tikz}
\usepackage{amsthm}

\usepackage{forest}
\usepackage{arydshln}
\usetikzlibrary{shapes.geometric,arrows.meta,positioning}
\newcommand{\colorboxinline}[1]{\protect\textcolor{#1}{\rule[0.05ex]{1em}{0.75em}}}

\definecolor{colFP}{HTML}{D55E00} % orange-red
\definecolor{colFN}{HTML}{56B4E9} % blue
\definecolor{colTP}{HTML}{009E73} % green

\usetikzlibrary{calc,positioning,arrows.meta,shapes.geometric,trees,matrix}
\definecolor{colFP}{HTML}{D55E00}
\definecolor{colFN}{HTML}{56B4E9}
\forestset{qtree/.style={for tree={parent anchor=south,child anchor=north,align=left,s sep=6mm,l sep=6mm}}}

\usepackage{enumitem}
\usepackage{IEEEtrantools} % provides \bstctlcite
\usepackage{adjustbox}

\setlist{nosep, leftmargin=14pt}

\title{Measuring and Aligning Abstraction in Vision-Language Models \\ with Medical Taxonomies}
\name{Ben Schaper$^{1,\ddagger}$\thanks{‡ Shared first authors}, Maxime Di Folco$^{1,2,3\ddagger}$, Bernhard Kainz$^{6,7}$, Julia A. Schnabel$^{1,2,4,5,*}$, Cosmin I. Bercea$^{1,4,*}$\thanks{* Co-senior authors}} 
\address{
    {$^1$} School of Computation, Information and Technology, Technical University of Munich, Germany\\
    {$^2$} Institute of Machine Learning in Biomedical Imaging, Helmholtz  Munich, Germany \\
    {$^3$} LTCI, Télécom Paris, Institut Polytechnique de Paris, France
    {$^4$} Munich Center for Machine Learning (MCML) \\
    {$^5$} School of Biomedical Engineering and Imaging Sciences, King's College London, UK\\
    {$^6$} Department of Artificial Intelligence in Biomedical Imaging, FAU Erlangen-Nuremberg, Germany \\ 
    {$^7$} Department of Computing, Imperial College London, UK
    }
\begin{document}
%\ninept
%
\bstctlcite{IEEEexample:BSTcontrol}
\maketitle
\begin{abstract}
Vision–Language Models (VLMs) show strong zero-shot performance for chest X-ray classification, but standard flat metrics fail to distinguish between clinically minor and severe errors. This work investigates how to quantify and mitigate abstraction errors by leveraging medical taxonomies. We benchmark several state-of-the-art VLMs using hierarchical metrics and introduce \textit{Catastrophic Abstraction Errors} to capture cross-branch mistakes. Our results reveal substantial misalignment of VLMs with clinical taxonomies despite high flat performance. To address this, we propose risk-constrained thresholding and taxonomy-aware fine-tuning with radial embeddings, which reduce severe abstraction errors to below 2\% while maintaining competitive performance. These findings highlight the importance of hierarchical evaluation and representation-level alignment for safer and more clinically meaningful deployment of VLMs.
\end{abstract}


% Introduction + overview figure (around 1 page)
\section{Introduction}
\label{sec:intro}

\begin{figure}[t]
    \centering
    \resizebox{1\linewidth}{!}{\input{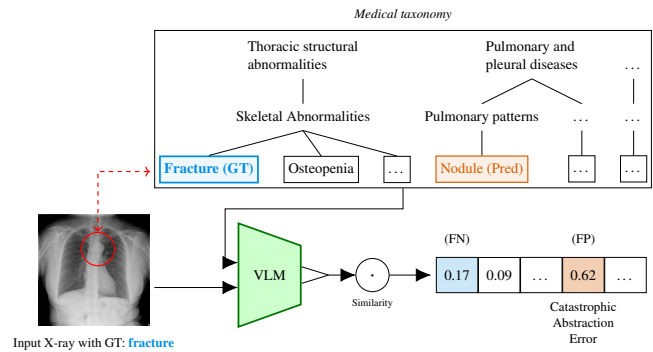}}
   \caption{Illustration of a Catastrophic Abstraction Error (CAE). A chest X-ray with ground truth label \textit{fracture} (FN, blue) is misclassified as \textit{nodule} (FP, orange). Since the prediction and ground truth lie in disjoint branches of the medical taxonomy (top), this cross-branch error constitutes a CAE.}
    \label{fig:overview}
\end{figure}

Vision–Language Models (VLMs) have rapidly advanced in recent years, driven by large-scale datasets and architectures capable of aligning images and text in a shared representation space \cite{Radford2021, Zhang2022}. A key strength of these models is their ability to perform zero-shot classification, where images can be categorised using textual descriptions of diseases or findings without task-specific training.  

In the medical domain, publicly available chest X-ray (CXR) datasets such as MIMIC-CXR \cite{Johnson2019}, and PadChest \cite{Castro2024} have enabled the adaptation of VLMs to radiology \cite{Wang2022,Eslami2023,Zhang2024}. However, most current evaluations rely on standard metrics such as accuracy or F1-score, which treat all errors equally. This uniform treatment obscures clinically critical mistakes: a misclassification between closely related findings (e.g., “cardiomegaly” vs. “enlarged cardiac silhouette”) is far less harmful than confusing findings from different categories (e.g., “rib fracture” vs. “pleural effusion”). Flat metrics, therefore, fail to reflect the clinical significance of errors.

Medical taxonomies such as ICD-10 \cite{Hirsch2016} and RadLex \cite{Wang2018} provide hierarchical structures that can capture semantic relationships between findings. Integrating these structures into evaluation frameworks allows for a more clinically meaningful assessment of VLMs by distinguishing between minor and catastrophic abstraction errors.  

% Contributions paragraph (can be modified)
This work investigates how to quantify and mitigate abstraction errors of VLMs in multi-label chest X-ray classification using hierarchical metrics and taxonomy-aware methods. The contributions of this paper are in threefolds: we (i) show that state-of-the-art VLMs remain misaligned with medical taxonomies under hierarchical metrics, (ii) we introduce a new hierarchical metric tailored to the multi-label setting to capture abstraction errors, and (iii) propose lightweight mitigation strategies, including a risk-constrained strategies for zero-shot multi-label classification and a taxonomy-aware fine-tuning, to reduce severe abstraction errors without compromising standard performance.

\section{Methods}
\label{sec:methods}

\subsection{Benchmarking Setup}

To test whether state-of-the-art (SOTA) VLMs are aligned with medical taxonomies, we benchmark their zero-shot performance in a multi-label setting using both flat and hierarchical metrics on PadChest \cite{Castro2024}. Here, flat metrics refer to classical evaluation measures that ignore the hierarchical structure, whereas hierarchical metrics explicitly account for it.

\textbf{Taxonomy Definition}:
The PadChest-GR dataset contains 4,555 frontal chest X-rays with bilingual, sentence-level annotations, including bounding boxes for localisable findings. These annotations are organised hierarchically, with labels grouped into clinically relevant categories that we adopt as a taxonomy. For this study, we extend the original taxonomy by introducing abstract 3 parent categories showed in Fig. \ref{fig:overview}, yielding a structured hierarchy of 117 nodes that capture both anatomical and pathological relationships between findings. The complete taxonomy is provided on GitHub\footnote{\url{https://github.com/compai-lab/2026-isbi-schaper}}.

\textbf{Models}: We benchmark a diverse set of SOTA VLMs % representing different pretraining strategies and scales, including 
from both general- and medical-domains: CLIP \cite{Radford2021}, PubMedCLIP \cite{Eslami2023}, BiomedCLIP \cite{Zhang2024}, MedCLIP \cite{Wang2022}, and MedSigLIP\footnote{\url{https://huggingface.co/google/medsiglip-448}}.

\textbf{Metrics}: For evaluation, we consider one flat baseline metric and three hierarchical metrics. The flat metric, macro-averaged $F_1$-score, serves as a standard, hierarchy-agnostic baseline. To capture clinically meaningful errors, we employ two established hierarchical metrics. The \textbf{Hierarchical Overlap Score (HOS)} \cite{Kiritchenko2005} expands the true and predicted label sets to include all ancestor nodes in the taxonomy. For example, a prediction of “\textit{Fracture}” is augmented with its parent “\textit{Skeletal abnormalities}” and grandparent “\textit{Thoracic structural abnormalities}.” Precision, recall, and F1 are then computed on these augmented sets to quantify semantic overlap. The \textbf{Hierarchical Distance Score (HDS)} \cite{Kosmopoulos2014} compute the $F_1$-score with partial credit to near-miss predictions based on the shortest-path distance between true and predicted labels in the taxonomy, penalising errors more heavily as the distance increases. %We also introduce a third hierarchical metric in the next section to specifically quantify catastrophic abstraction errors. 
%Together, these metrics enable assessment of both overall classification performance and alignment with the hierarchical structure of medical knowledge. 
Figure 2 illustrates three representative classification scenarios that highlight the complementary behaviour of taxonomy-aware metrics. In (a), a prediction that would score zero under flat $F_1$ receives partial credit from both hierarchical metrics due to taxonomic proximity of errors. In (b), the Overlap Score is relatively high as it rewards taxonomic consistency despite several false positives, whereas the Distance Score penalises them more strongly. Conversely, (c) shows a case where the Distance Score remains high by rewarding semantic proximity of predictions, while the Overlap Score penalises out-of-branch errors more severely.

\begin{figure}[h]
\centering

% Define colors (these are for the fills)
\definecolor{truepositive}{RGB}{34, 139, 34}
\definecolor{falsepositive}{RGB}{255, 140, 0}
\definecolor{falsenegative}{RGB}{70, 130, 180}

% Define styles
\tikzset{
    % Base style for all nodes: a circle with a light fill and black text
    word/.style={
        font=\scriptsize,  % Smaller font
        circle,            % Draw as a circle
        draw=none,         % No border
        text=black,        % Set text color to black
        minimum size=0.45cm, % Circle size
        inner sep=1pt,     % Padding
        align=center
    },
    % Styles for each classification type, inheriting from 'word'
    % These now only change the fill color with light opacity
    tp/.style={word, fill=truepositive!20}, % True Positive (light green fill)
    fp/.style={word, fill=falsepositive!20}, % False Positive (light orange fill)
    fn/.style={word, fill=falsenegative!20}, % False Negative (light blue fill)
    thinline/.style={gray, dashed, line width=0.3pt},
    thickline/.style={black, line width=0.8pt}
}

% --- COORDINATES ADJUSTED FOR TIGHTER SPACING ---
\begin{tikzpicture}[scale=0.95]

% Example (a)
\node at (-1, 0) {\small (a)};

% A1 group
\node[fp] at (0.4, 0) {A1.1};
\node[fp] at (1.15, 0) {A1.6};
\node[fn] at (1.9, 0) {A1.7};

% Thick divider between A and B
\draw[thickline] (2.3, -0.25) -- (2.3, 0.25);

% B3 group
\node[fn] at (2.7, 0) {B3.1};
\node[fp] at (3.45, 0) {B3.2};

% Example (b)
\node at (-1, -1) {\small (b)};

% A1 group
\node[fn] at (0.4, -1) {A1.5};
\node[fp] at (1.15, -1) {A1.6};

% Thick divider between A and B
\draw[thickline] (1.55, -1.25) -- (1.55, -0.75);

% B1 group
\node[fn] at (1.95, -1) {B1.4};

% Example (c)
\node at (-1, -2) {\small (c)};

% First line - A1 group
\node[fp] at (0.4, -2) {A1.1};
\node[fp] at (1.15, -2) {A1.2};
\node[fp] at (1.9, -2) {A1.3};
\node[tp] at (2.65, -2) {A1.4};
\node[fp] at (3.4, -2) {A1.5};

% Continue on second line
\node[tp] at (0.4, -2.8) {A1.6};
\node[fp] at (1.15, -2.8) {A1.7};

% Thin divider between A1 and A2
\draw[thinline] (1.55, -3.05) -- (1.55, -2.55);

% A2 group
\node[fp] at (1.95, -2.8) {A2.2};

% Thick divider between A and B
\draw[thickline] (2.35, -3.05) -- (2.35, -2.55);

% B1 group
\node[fp] at (2.75, -2.8) {B1.6};

% Thin divider between B1 and B3
\draw[thinline] (3.15, -3.05) -- (3.15, -2.55);

% B3 group
\node[fp] at (3.55, -2.8) {B3.2};

\end{tikzpicture}
\caption{Three classification examples illustrating the complementary behaviour of taxonomy-aware metrics (full taxonomy omitted). \colorboxinline{truepositive!20}: true positives; \colorboxinline{falsepositive!20}: false positives; \colorboxinline{falsenegative!20}: false negatives. Vertical bars separate taxonomic branches: thick bars for main branches, thin bars for sub-branches. %(a) A case where the flat $F_1$ score is zero, but hierarchical metrics (HOS \& HDS) provide partial credit for taxonomically proximate errors. (b) The Hierarchical Overlap Score (HOS) rewards branch consistency, while the Hierarchical Distance Score (HDS) penalises the false positives more. (c) HDS rewards the semantic proximity of predictions, whereas HOS strongly penalises the out-of-branch predictions.
}
\label{fig:metrics}
\end{figure}
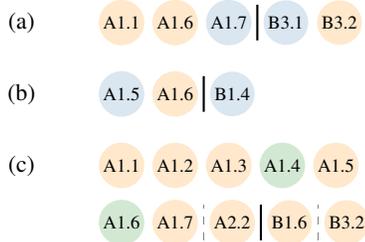

\begin{table*}[t]
	\centering
        \caption{Multi-label classification performance of VLMs in a zero-shot setting using performance-based thresholds (Risk $\times$) or the proposed risk-constrained thresholds (Risk $\checkmark$), compared against the proposed taxonomy-aware fine-tuning. Mean and bootstrapped 95$\%$ confidence intervals are shown. All metrics are in $\%$. Bold indicates best, underline indicates second-best.}
	\begin{scriptsize}
 \begin{tabular}{lcccc}
     \toprule
     \textbf{Model} & $\mathbf{F_1}$ $\uparrow$& \textbf{HDS} $\uparrow$ & \textbf{HOS} $\uparrow$ & \textbf{CAE} $\downarrow$  \\
     \hline
     \multicolumn{5}{c}{\textit{Performance-based threshold (Risk $\times$)}} \\
     \hline
     \rowcolor{gray!20} MedSigLIP & \textbf{22.35} (20.23, 24.26)  & \textbf{21.49} (18.92, 23.90)  & \textbf{43.23} (41.39, 45.02)  & 19.5 \\
     BiomedCLIP & 15.36 (13.23, 17.52)  & 8.71 (6.04, 11.43)  & \underline{40.30} (38.53, 42.05)  & 11.6 \\
     \rowcolor{gray!20} CLIP Base & 11.51 (10.24, 12.83)  & 3.43 (1.84, 5.35)  & 33.27 (32.18, 34.49)  & \underline{0.5} \\
     PubMedCLIP & 9.68 (8.49, 10.82)    & 2.48 (0.56, 4.27)  & 37.64 (36.28, 38.92) & 1.5 \\
     \rowcolor{gray!20} MedCLIP & 14.04 (13.41, 14.69)  & 0.03 (0.00, 0.07) & 24.52 (23.68, 25.45) & \textbf{0.2} \\
     \midrule
     \multicolumn{5}{c}{\textit{Risk-constrained threshold (Risk $\checkmark$)}}\\
     \hline
     \rowcolor{gray!20} MedSigLIP & {17.93} (16.69, 19.19)    & 8.24 (7.21, 9.58)             & 31.65 (30.55, 32.78) & 0.9 \\
     BiomedCLIP & 14.75 (13.35, 16.18)          & 6.98 (5.07, 9.17)             & {37.57} (36.09, 39.05) & 2.4 \\
     \rowcolor{gray!20} CLIP Base & 10.14 (9.13, 11.16)           & 2.32 (1.05, 3.45)             & 34.56 (33.34, 35.76) & 1.5 \\
     PubMedCLIP & 10.09 (8.78, 11.29) & 3.61 (1.71, 5.37) & 38.55 (37.19, 39.95) & 2.2 \\
     \rowcolor{gray!20} MedCLIP & 13.99 (13.33, 14.65)          & 0.03 ($-$0.03, 0.10)          & 24.45 (23.62, 25.36) & {0.7} \\
\hdashline  % requires \usepackage{arydshln}
CLIP Base + TAF (\textit{Ours}) & \underline{21.17} (19.74, 22.51) & \underline{15.01} (13.53, 16.80) & 33.69 (32.56, 34.81) & 1.6 \\
\bottomrule
 \end{tabular}
\end{scriptsize}

	\label{tab:main}
\end{table*}

\subsection{Quantifying Catastrophic Abstraction Errors}
% Illustrated in the overview figure normally

We introduce the concept of a Catastrophic Abstraction Error (CAE) to capture errors that are clinically critical in hierarchical classification. For example, misclassifying a \textit{fracture} as a \textit{nodule} (Fig.~\ref{fig:overview}) constitutes a CAE, whereas confusing it with \textit{osteopenia} is less severe. Formally, a CAE occurs when the predicted and true label sets lie in disjoint subtrees that share only the root as a common ancestor. In Fig.~\ref{fig:overview}, \textit{nodule} belongs to the \textit{Pulmonary and pleural diseases} branch, whereas \textit{fracture} belongs to \textit{Thoracic structural abnormalities}. Such errors represent a complete semantic mismatch and are not captured by flat metrics such as $F_1$, which weight all false positives equally regardless of semantic distance.

\subsection{Threshold Selection for Zero-Shot classification}

We propose a novel \textit{risk-constrained} threshold selection for zero-shot classification to limit the allowable CAE rate. In contrast to the normal procedure of simply maximising $F_1$ on the validation set (denoted \textit{performance-based} in the paper), we add a safety constraint by selecting the threshold $\delta_{\text{safety}}$ by finding the best possible $F_1$-score while making sure the CAE rate stays below a predefined maximum, $\tau$. A key observation is that the CAE rate changes monotonically with the prediction threshold. By lowering the threshold, the model is more likely to make a correct prediction from the same branch as the ground truth, which reduces the chance of a catastrophic error. Our risk-constrained approach exploits this by choosing the threshold that achieves the best $F_1$ score without exceeding the pre-defined CAE limit of $\tau$ $\leq$1\%.

\subsection{Taxonomy-Aware Fine-Tuning}

To reduce severe CAEs, we evaluate two complementary fine-tuning strategies, each designed to improve alignment with the PadChest-GR taxonomy. First, \textbf{SigLIP Fine-Tuning}, an in-domain adaptation without hierarchy, fine-tune a general-domain model using SigLIP’s pairwise sigmoid loss \cite{Zhai2023} on the PadChest-GR \cite{Castro2024}, which includes finding sentences and labels. Freeze the vision and text encoders, updating only the projection layers to control efficiency and prevent overfitting, testing whether in-domain adaptation alone can reduce catastrophic errors. Secondly, \textbf{Radial Embedding (RE)} Fine-Tuning \cite{Alper2024}, an explicit hierarchical encoding strategy, leverages an RE loss to adjust the concept embeddings’ radial distances from the hierarchy’s root, explicitly separating incompatible branches. Root and branches are defined by adapting the HierarCaps framework \cite{Alper2024} to PadChest-GR by constructing, for each finding, a “positive chain” following its root-to-leaf path and a “negative chain” by substituting a mutually exclusive sibling. For instance, the finding "\textit{Signs of air trapping}" is linked to the following positive chain: "\textit{Pulmonary and Pleural Diseases}" (abstract) $\rightarrow$ "\textit{Pulmonary Patterns}" $\rightarrow$ "\textit{Hyperinflated lung}" $\rightarrow$ "\textit{Signs of air trapping} (specific)". Negative chains are created by substituting mutually exclusive siblings at each level of abstraction, e.g., substituting "\textit{Pulmonary and Pleural Diseases}" with "\textit{Thoracic Structural Abnormalities}". We freeze the vision encoder to preserve the original multimodal alignment. %By comparing these two approaches and a combination of both, we can determine whether general-domain adaptation or explicit hierarchical encoding is more effective at reducing catastrophic errors.

% \subsection{Implementation Details}

% All zero-shot results use PadChest-GR frontal X-rays at 224×224 (448×448 for MedSigLIP). Text features are averaged over five sampled finding sentences per label. CLIP ViT-B/32 fine-tuning uses multi-positive SigLIP loss, updating projection layers, logit scale, and SigLIP parameters (AdamW, lr=1e-4, weight decay 0.01, batch size 128). Early stopping triggers when validation distance $F_1$ fails to improve for 10 checks. Radial embedding fine-tuning starts from this checkpoint, freezes the vision encoder, and updates text encoder and projection for one epoch (lr=1e-7, batch size 8). Hardware: NVIDIA RTX A6000. Additional implementation details and code are available at our repository.

%Evaluation follows three steps: (1) (2) threshold selection on the validation set, either performance-based—defined as maximising a joint score that combines normalised flat, overlap, and distance $F_\beta$ metrics with equal weights—or CAE-constrained with a limit of $\leq1\%$; and (3) binarization to compute flat and hierarchical $F_1$ scores.
%(1) inference with a script adapted to the three VLMs implementations (\texttt{transformers} \cite{Wolf2020} for CLIP and SigLIP, \texttt{open\_clip} \cite{Ilharco2021}, and the MedCLIP-specific implementation \cite{Wang2022});

%together with a chance–corrected distance metric, with 95\% confidence intervals obtained by bootstrap resampling at the image level. 

\section{Experiments and results}
\label{sec:exp}

\subsection{Taxonomy-Aware Fine-Tuning}  
Table~\ref{tab:main} reports the zero-shot multi-label classification results. In the standard setting (top table), MedSigLIP achieves the highest performance across $F_1$, HDS, and HOS metrics, followed by BiomedCLIP. However, both models exhibit CAE rates above 10\%, indicating misalignment with medical taxonomies despite their strong overall performance. Models with lower CAE rates achieve much lower $F_1$ scores, highlighting a trade-off between predictive accuracy and catastrophic error reduction. To address the misalignment observed in the zero-shot setting, we apply the proposed \textit{risk-constrained threshold} (bottom table), which directly targets catastrophic errors by adjusting the decision threshold to keep CAE rates below a predefined limit. This reduces CAE to below 2.4\%, effectively eliminating most cross-branch misclassifications highlighted earlier. While this comes at the cost of a moderate drop in $F_1$, the improvement in semantic consistency demonstrates that threshold optimisation is an effective first step toward aligning VLM predictions with medical taxonomies.
%\subsection{Taxonomy-Aware Fine-Tuning}  
%While threshold optimisation reduces the frequency of catastrophic errors,
However, it cannot recover the performance lost. To address this, we proposed a \textit{taxonomy-aware fine-tuning} strategy, which explicitly aligns the embedding space with the hierarchical structure of medical findings (CLIP Base + TAF). As shown in Table~\ref{tab:main}, our approach achieves competitive $F_1$ performance compared to MedSigLIP in the performance-based setting, while keeping CAE rates low — a combination that neither zero-shot models nor risk-constrained thresholding alone can offer. The ablation study in Table~\ref{tab:ablation} further highlights the complementary roles of each component. We find that \textit{performance-based} thresholding yields the highest $F_1$ but allows many catastrophic errors to persist, even with fine-tuning. Under the \textit{risk-constrained} threshold, CAE rates are successfully controlled, and the combination of SigLIP fine-tuning with radial embedding yields the best balance, maximising performance while minimising severe cross-branch errors. Also, radial embedding fine-tuning alone does not provide improvement over the baseline, indicating that hierarchical supervision without domain adaptation (SigLIP fine-tuning) is insufficient in this setting.

\begin{table}[t]
	\centering
    \caption{Ablation Study of the Fine-Tuning strategy using performance-based (Perf.) or risk-constrained (Risk) threshold. RE: Radial Embedding. Bold indicates best, underline indicates second-best. \colorboxinline{BrickRed!20} highlights CAE-critical values; \colorboxinline{ForestGreen!20} indicates best results with CAE below 5$\%$.}
	\centering
\begin{adjustbox}{width=0.65\linewidth}
\begin{tabular}{c|c|c|cccc}
\toprule
    \multicolumn{1}{c}{} & \multicolumn{1}{c}{\textbf{SigLIP}} & \multicolumn{1}{c}{\textbf{RE}} & \textbf{$F_1$}  $\uparrow$ & \textbf{HDS $\uparrow$} & \textbf{HOS}  $\uparrow$ & \textbf{CAE} $\downarrow$ \\
    
    \midrule
           \multirow{4}{*}{\rotatebox{90}{Perf.}}  &  &     & 11.51  & 3.43    & 33.27  & \textbf{0.50} \\
               &  & \checkmark & 10.80   & 1.19  & 30.73  & \underline{1.10} \\
    & \checkmark  &            & \cellcolor{BrickRed!20}\textbf{24.38}  & \cellcolor{BrickRed!20}\textbf{19.60} & \cellcolor{BrickRed!20}\textbf{39.02} & \cellcolor{BrickRed!20}{35.60} \\
    & \checkmark  & \checkmark & \cellcolor{BrickRed!20}\underline{21.79}  & \cellcolor{BrickRed!20}\underline{17.99}  & \cellcolor{BrickRed!20}\underline{37.30}  & \cellcolor{BrickRed!20}{6.40} \\
    
    \midrule
               \multirow{4}{*}{\rotatebox{90}{Risk}} & & & 10.14   & 2.32    & \cellcolor{ForestGreen!20}{34.56}    & {1.50} \\
               &  & \checkmark & 10.87  & 1.37 & 31.06   & 1.80 \\
    & \checkmark  &           & {17.33}  & {5.71}  & 30.26   & 3.50 \\
    & \checkmark  & \checkmark & \cellcolor{ForestGreen!20}{21.17}  & \cellcolor{ForestGreen!20}{15.01} & {33.69}  & {1.60} \\
\bottomrule
\end{tabular}
\end{adjustbox}

	\label{tab:ablation}
\end{table}

\subsection{Representation Alignment with Taxonomy}  

\begin{table}[h]
	\centering
    \caption{Kendall’s $tau$ rank correlation between model predictions and PadChest-GR taxonomy
ordering on the test set. Mean ± SD across all root-to-leaf paths. Higher indicates
stronger preservation of taxonomic abstraction order.}
	\begin{adjustbox}{width=\linewidth}
\begin{tabular}{lcccccc}
\toprule
\textbf{Model} & \textit{Ours} & PubMedCLIP & CLIP Base & MedCLIP & MedSigLIP & BiomedCLIP \\
\midrule
\textbf{Kendall's $\tau$} & 
\textbf{0.86} $\pm$ & 
\underline{0.72} $\pm$  & 
0.25 $\pm$  & 
0.19 $\pm$  & 
$-$0.17 $\pm$  & 
$-$0.69 $\pm$ \\
SD & 0.27 & 0.38 & 0.48 & 0.43 & 0.55 & 0.51 \\
\bottomrule
\end{tabular}
\end{adjustbox}

	\label{tab:kendall}
\end{table}

Finally, we analyse whether the improved performance corresponds to better alignment between the model’s representation space and the medical taxonomy. Using Kendall’s $\tau$ rank correlation \cite{Alper2024}, which measures whether model representations preserve the semantic ordering from general to specific concepts, we find that our hierarchically fine-tuned model achieves $\tau=0.86$, a substantial improvement over the CLIP Base baseline ($\tau=0.25$), confirming that radial embedding fine-tuning effectively encodes hierarchical relationships. In contrast, several medical-domain VLMs exhibit poor or even negative alignment, with BiomedCLIP showing strong anti-correlation ($\tau \approx -0.69$). These results suggest that improved taxonomy alignment translates into lower CAE rates, linking the quantitative performance gains to meaningful changes in representation geometry.

\section{Conclusion}
\label{sec:conclusion}

This work investigated how to quantify and mitigate abstraction errors of VLMs in multi-label CXR classification using hierarchical metrics and taxonomy-aware methods. We first showed that SOTA VLMs remain misaligned with medical taxonomies under hierarchical evaluation, with standard flat metrics obscuring severe cross-branch errors. To capture these clinically critical mistakes, we introduced Catastrophic Abstraction Errors (CAEs), a hierarchical metric tailored to the multi-label setting. Finally, we proposed lightweight mitigation strategies, including risk-constrained thresholding for zero-shot classification and taxonomy-aware fine-tuning with radial embeddings, which substantially reduce severe abstraction errors without compromising overall performance. Our results show that hierarchical evaluation, risk-aware inference, and representation-level alignment provide a path toward safer and more clinically reliable multimodal models.

% \section{Compliance with ethical standards}
% \label{sec:ethics}

% This research study was conducted retrospectively using human subject data made available in open access by \cite{Castro2024}. Ethical approval was not required as confirmed by the license attached with the open-access data.

% \section{Conflict of Interest}

% No funding was received for conducting this study. The authors have no relevant financial or non-financial interests to disclose.

\bibliographystyle{IEEEtran}
\bibliography{strings,refs}

%\section{Acknowledgments}
%\label{sec:acknowledgments}
%
%IEEE-ISBI supports the disclosure of financial support for the project
%as well as any financial and personal relationships of the author that
%could create even the appearance of bias in the published work. The
%authors must disclose any agency or individual that provided financial
%support for the work as well as any personal or financial or
%employment relationship between any author and the sources of
%financial support for the work.
%
%Other types of acknowledgements can also be listed in this section.
%
%Reporting on real or potential conflicts of interests, or the absence
%thereof, is required in the paper. Authors are responsible for
%correctness of the statements provided in the manuscript. Examples of
%appropriate statements include:
%\begin{itemize}
%  \item ``No funding was received for conducting this study. The
%    authors have no relevant financial or non-financial interests to
%    disclose.'' 
%  \item ``This work was supported by […] (Grant numbers) and
%    […]. Author X has served on advisory boards for Company Y.'' 
%  \item ``Author X is partially funded by Y. Author Z is a Founder and
%    Director for Company C.''
%\end{itemize}


\end{document}